\documentclass[letterpaper, 10 pt, conference]{ieeeconf}  % Comment this line out if you need a4paper

\usepackage{amsthm}
\usepackage{amsmath}
\usepackage{graphicx}
\usepackage{cite}
\usepackage[utf8]{inputenc}
\usepackage{picinpar}
\usepackage{url}
\usepackage{flushend}
\usepackage{colortbl}
\usepackage{soul}
\usepackage{multirow}
\usepackage{pifont}
\usepackage{color}
\usepackage{alltt}
\usepackage[hidelinks]{hyperref}
\usepackage{enumerate}
\usepackage{siunitx}
\usepackage{breakurl}
\usepackage{epstopdf}
\usepackage{pbox}
\usepackage{amsfonts,amssymb}
\usepackage{mathrsfs}

\renewcommand{\Re}{\operatorname{Re}}
\renewcommand{\Im}{\operatorname{Im}}

\IEEEoverridecommandlockouts                              % This command is only needed if 
\title{\LARGE \bf
Conformal Navigation Transformations with Application to Robot Navigation in Complex Workspaces
}

\author{Li Fan$^{1}$, Jianchang Liu$^{1}$, Wenle Zhang$^{2}$ and Peng Xu$^{1}$% <-this % stops a space
\thanks{$^{1}$Li Fan, Jianchang Liu and Peng Xu are with the State Key Laboratory of Synthetical Automation for Process Industries, Northeastern University, Shenyang 110819, China, and also with the College of Information Science and Engineering, Northeastern University, Shenyang 110819, China 
        {\tt\small lifanneu@yeah.net;liujianchang@ise.neu.edu.cn;
        xpps0827@gmail.com}}%
\thanks{$^{2}$Wenle Zhang is with the College of Engineering, Ocean University of China, Qingdao 266100, China
        {\tt\small zhangwenle@ecust.edu.cn}}%
}

\begin{document}

\maketitle
%\pagestyle{empty}  
%\thispagestyle{empty}  
%%%%%%%%%%%%%%%%%%%%%%%%%%%%%%%%%%%%%%%%%%%%%%%%%%%%%%%%%%%%%%%%%%%%%%%%%%%%%%%%
\begin{abstract}

Navigation functions provide both path and motion planning, which can be used to ensure obstacle avoidance and convergence in the sphere world. When dealing with complex and realistic scenarios, constructing a transformation to the sphere world is essential and, at the same time, challenging. %When dealing with complex and realistic scenarios, constructing a transformation to the sphere world is both essential and challenging.
%Navigation functions provide both path and motion planning, which can be used to achieve collision-free navigation of mobile robots.
%The navigation function method can simultaneously solve the path and motion planning problems of mobile robots. When dealing with complex and realistic scenarios, it mostly relies on a diffeomorphism transformation to the sphere world.
%The navigation function method can simultaneously solve the path and motion planning problems of mobile robots, and it mostly relies on a diffeomorphism transformation to the sphere world when handling complex and realistic scenarios.
This work proposes a novel transformation termed the conformal navigation transformation to achieve collision-free navigation of a robot in a workspace populated with obstacles of arbitrary shapes. The properties of the conformal navigation transformation, including uniqueness, invariance of navigation properties, and no angular deformation, are investigated, which contribute to the solution of the robot navigation problem in complex environments. %which facilitates the navigation of robots in complex environments. 
Based on navigation functions and the proposed transformation, feedback controllers are derived for the automatic guidance and motion control of kinematic and dynamic mobile robots. Moreover, an iterative method is proposed to construct the conformal navigation transformation in a multiply-connected workspace, which transforms the multiply-connected problem into multiple simply-connected problems to achieve fast convergence. % an iterative construction method
In addition to the analytic guarantees, simulation studies verify the effectiveness of the proposed methodology in workspaces with non-trivial obstacles.
%The theoretical results provided are supported by analytical proofs. Simulation studies are conducted to demonstrate the effectiveness of the methodology.%control scheme

%The navigation function method can simultaneously solve the path and motion planning problems of mobile robots, and it mostly relies on a diffeomorphism transformation to the sphere world when handling complex and realistic scenarios.This work proposes a novel transformation termed the conformal navigation transformation to achieve collision-free navigation of a robot in a workspace populated with obstacles of arbitrary shapes. The properties of the conformal navigation transformation, including uniqueness, invariance of navigation properties and no angular deformation, are investigated in this work, which facilitates the navigation of robots in complex environments. Based on navigation functions and conformal navigation transformations, feedback controllers are derived for automatic guidance and motion control of kinematic and dynamic mobile robots. Moreover, an iterative construction method for the conformal navigation transformation on a multi-connected workspace is proposed, which transforms the multi-connected problem into multiple single-connected problems to achieve fast convergence. The theoretical results provided are supported by analytical proofs. Simulation studies are conducted to demonstrate the effectiveness of the proposed control scheme.
\end{abstract}

%\begin{keywords}
%Motion and path planning, collision avoidance, navigation function, conformal map
%\end{keywords}

%%%%%%%%%%%%%%%%%%%%%%%%%%%%%%%%%%%%%%%%%%%%%%%%%%%%%%%%%%%%%%%%%%%%%%%%%%%%%%%%
\section{Introduction} %INTRODUCTION

The autonomous navigation of robots in cluttered environments is an actively studied topic of robotics research. Control laws based on artificial potential fields (APFs) constitute a widely researched tool for solving the robot navigation problem due to their intuitive design and ability to simultaneously solve path planning and motion planning subproblems \cite{1,2}. On the other hand, this control method suffers from the presence of local minima and the inability to cope well with non-convex obstacles.%this control method suffers from the problem of local minima, and cannot cope well with non-convex obstacles.

The navigation function \cite{Rimon1992,6,7} method proposed by Rimon and Koditschek is an efficient method for constructing a class of local minima free APFs in the sphere world, which can be extended to environments with star-shaped obstacles by constructing a diffeomorphism. The negative gradient of the navigation function can be served as a control input to guarantee collision-free motion of the kinematic mobile robot and convergence to the goal point from almost all initial points. A large number of successful applications have appeared in the literature, such as multi-robot systems \cite{Loizou2008,11,12}, highly dynamic or partially known environments \cite{Li2019,Loizou2022}%\cite{Li2019,Loizou2022,14}
, uncertain dynamic systems \cite{15}, and constrained stabilization problems \cite{16}. In a recent work \cite{17}, the construction of navigation functions was extended to non-spherical convex obstacles; however, the construction of navigation functions in geometrically complex spaces is still a challenging problem. %however constructing navigation functions in non-trivial spaces is still a challenging problem.

Transforming geometrically complex spaces to geometrically simple spaces is an important and elegant approach to implementing %applying
navigation functions in complex workspaces. Techniques based on diffeomorphism and navigation transformation are a few of the promising methods found in the literature to achieve the above objective. In \cite{6}, a family of analytic diffeomorphisms was constructed for mapping any star world %$\footnote{A star set is a set that possesses a point from which all the rays cross the boundary only once. A star world is formed by removing from the interior of a large star set a finite number of non-overlapping smaller star sets.}$ 
to an appropriate sphere world, requiring appropriate tuning of a certain parameter. To cope with the complexity of the workspace, the authors in \cite{18} proposed a diffeomorphism based on the harmonic map, which maps the workspace to a punctured disc by solving multiple boundary problems. By applying a two-step navigation transformation, the author in \cite{19} achieved a tuning-free solution by pulling back a trivial solution in the point world to the initial star world. In a subsequent work \cite{Nicolas2020}, the authors proposed a single-step navigation transformation candidate that directly maps a known star world to a point world.

The transformation candidates in most of the aforementioned works are only suitable for star worlds rather than arbitrary workspaces, which hinders the application of navigation functions in realistic environments. %Most of the aforementioned works depend on a transformation from the workspace to the sphere world, but the transformation candidates are mainly suitable for the star world rather than an arbitrary workspace. %applicable to
In this work, we address the problem of navigating kinematic and dynamic mobile robot systems in a workspace with arbitrary connectedness and shape by constructing a novel transformation termed the conformal navigation transformation. Unlike the transformations mentioned above, the conformal navigation transformation requires neither tuning a parameter nor decomposing the workspace into trees of stars, and it is unique in a given workspace. Besides, it can pull back the navigation function in the sphere world to the initial workspace, as well as the trivial solution in the sphere world determined by the vector field, so that the navigation problem in the workspace can be solved elegantly. Further, we present a fast iterative method for constructing the proposed transformations in any complex workspace. %In addition, we present a method to obtain conformal navigation transformations in multi-connected workspaces by iterating over the conformal navigation transformations in single-connected workspaces, while also reducing the computational load required to deal with complex workspaces.
To the best of the authors' knowledge, this is the first work to provide a 
tuning-free transformation candidate that maps any workspace with non-star-shaped obstacles to a sphere world. The innovations and contributions of this work are summarized in the following.

\begin{enumerate}
	\item The conformal navigation transformation is proposed to solve the navigation problem for a workspace with arbitrary shape and countably many boundary components. The proposed transformation has no angular distortion, no tuning, and is unique for a given workspace.
	%A novel transformation called the conformal navigation transformation is proposed for a workspace with arbitrary shape and at most countably infinite many boundary components. The proposed transformation has no angular distortion, no tuning and is unique for a given workspace.
	%Introduction of the conformal navigation transformation, analysis of its properties suitable for navigation problems in complex workspaces.
	\item The provably correct feedback control laws for kinematic and dynamic mobile robots are designed to achieve convergence from almost all initial points to the goal point.
	%Space coordinate transformations with no angular deformation, no tuning, and unique for a given workspace.
	\item An iterative method is presented to rapidly construct the conformal navigation transformation, which is suitable for non-star-shaped obstacles.
	%Fast iterative method for conformal navigation transformations applicable to arbitrary obstacle shapes.
\end{enumerate}

The rest of this paper is organized as follows. Section \uppercase\expandafter{\romannumeral2} introduces the necessary preliminaries. Section \uppercase\expandafter{\romannumeral3} gives the definition of the conformal navigation transformation and analyzes its properties. Section \uppercase\expandafter{\romannumeral4} presents two mobile robot controllers designed using conformal navigation transformations and navigation functions. Section \uppercase\expandafter{\romannumeral5} proposes the construction method of conformal navigation transformations. Section \uppercase\expandafter{\romannumeral6} provides simulation studies to demonstrate the effectiveness of the proposed control scheme. Section \uppercase\expandafter{\romannumeral7} concludes this paper.

\section{Preliminaries}
This section introduces sphere worlds and workspaces, the navigation function, the mobile robot models, and the conformal map.

\subsection{Sphere Worlds and Workspaces}
The $n$-dimensional sphere world $\mathscr{M}^n$ as defined in \cite{Rimon1992} is a compact connected subset of ${\mathbb{R}^n}$ whose boundary is formed by the disjoint union of an external $(n - 1)$-sphere ${\mathscr{\tilde{O}}_0}$ and other internal $(n - 1)$-spheres ${\mathscr{\tilde{O}}_i}$ that define obstacles in ${\mathbb{R}^n}$. Each internal sphere obstacle ${\mathscr{\tilde{O}}_i}$ is the interior of the $(n - 1)$-sphere, which is implicitly defined as $\mathscr{\tilde{O}}_i=\{q\in\mathbb{R}^n:\|q-q_i\|^2<\rho_i^2\},\ i\in\{1,...,M\}.$ For simplicity, the exterior of the external $(n - 1)$-sphere is refer to as the zeroth sphere obstacle, denoted as ${\mathscr{\tilde{O}}_0}$. In this regard, the $n$-dimensional sphere world with $M$ internal obstacles $\mathscr{M}^n$ is represented as:\begin{equation}\begin{split}\label{eq1}\mathscr{M}^n=\{&q\in\mathbb{R}^n:-\|q-q_0\|^2\geqslant-\rho_0^2,\\&\|q-q_1\|^2\geqslant\rho_1^2,...,\|q-q_M\|^2\geqslant\rho_M^2\}.\end{split}\end{equation}

Clearly, the sphere world is geometrically simple and can be served as a topological model of geometrically complex spaces. We will restrict our attention to the geometrically complex robot workspace, defined as follows:

\textit{Definition 1:} The $n$-dimensional robot workspace ${\mathscr{N}^n}\subset\mathbb{R}^{n}$ is a connected and compact $n$-dimensional manifold with boundary %in ${\mathbb{R}^{m\geqslant n}}$
such that ${\mathscr{N}^n}$ is homeomorphic to $\mathscr{M}^n$.

For a finite $M\in{\mathbb{N}}$, %$M\in{\mathbb{Z}^{+}}$ 
the internal obstacle ${\mathscr{O}_i}$ of ${\mathscr{N}^n}$ is the interior of a connected and compact $n$-dimensional subset of ${\mathbb{R}^n}$ such that $\mathscr{O}_i\cap\mathscr{O}_j=\varnothing,\ i\ne j,\ i,j\in\{1,...,M\}.$ It is convenient to regard the unbounded component of the workspace's complement as the zeroth obstacle ${\mathscr{O}_0}$. Each obstacle can be represented by an obstacle function $\beta_i:{\mathbb{R}}^n\to\mathbb{R}$ in the following form: 
\begin{equation}\label{eq1}\mathscr{O}_i=\{q\in\mathbb{R}^n:\beta_i(q)<0\},\ i\in\{0,...,M\}.\end{equation}

For the obstacle function, zero is not a critical value, and the zero level set of $\beta_i$ defines the obstacle's boundary: $\partial\mathscr{O}_i=\{q\in\mathbb{R}^n:\beta_i(q)=0\},\ i\in\{0,...,M\}$. Moreover, according to the implicit function theorem, the boundary of an obstacle is an $(n-1)$-dimensional  submanifold of ${\mathscr{N}^n}$.
Then, the $n$-dimensional workspace with $M$ internal obstacles ${\mathscr{N}^n}$ is represented as: \begin{equation}\label{eq2}\mathscr{N}^n=\{q\in\mathbb{R}^n:\beta_0(q)\geqslant0,...,\beta_M(q)\geqslant0\}.\end{equation}

\subsection{Navigation Function}
The navigation function is a well-established technique for robot navigation in spaces populated with obstacles. For almost all initial states, the control law generated by the navigation function defines asymptotically stable closed-loop robot systems with collision-free trajectories.

\textit{Definition 2 \cite{Rimon1992}:} Let $\mathscr{F}\subset\mathbb{R}^n$ be a compact connected $n$-dimensional analytic manifold with boundary, and let $x_d$ be a goal point in the interior of $\mathscr{F}$. A map $\phi:\mathscr{F}\to[0,1]$, is a navigation function if it is:
\begin{enumerate}
	\item Smooth on $\mathscr{F}$: at least twice continuously differentiable on $\mathscr{F}$.
	\item Admissible on $\mathscr{F}$: uniformly maximal on $\partial\mathscr{F}$.
	\item Polar at $x_d$: has a unique minimum at $x_d$.
	\item Morse on $\mathscr{F}$: has only non-degenerate critical points on $\mathscr{F}$.
\end{enumerate}

It can be proved that a smooth navigation function exists for every smooth compact connected manifold with boundary and any interior goal point. 
The Koditschek-Rimon navigation function $\phi_{kr}=\frac{\|q-q_d\|}{{(\|q-q_d\|^k+\beta)}^{1/k}}$, is a navigation function constructed in the sphere world, so long as $k$ exceeds some threshold $k_{min}$, where $\beta=\prod_{j=0}^M\beta_i$ is the aggregate obstacle function.

Unfortunately, there is not yet an efficient way to construct a navigation function for the general case. Based on the conformal navigation transformation proposed in Section \uppercase\expandafter{\romannumeral3}, this paper presents a geometric approach to constructing navigation functions in geometrically complex workspaces.

\subsection{Mobile Robot Models}
This work considers two types of mobile robots that navigate in a $2$-dimensional workspace $\mathscr{N}^2\subset\mathbb{R}^2$.

The motion of the first type of mobile robot is described by the trivial kinematic integrator model:
\begin{equation}\dot{x}(t)=u(t),\label{kinematic}\end{equation}
where $x\in\mathbb{R}^2$ is the robot position and $u\in\mathbb{R}^2$ is the corresponding control input vector. The second model is the second-order dynamic integrator for a mobile robot of mass $m$:\begin{equation} m\cdot\ddot{x}(t)=\tau(t),\label{dynamic}\end{equation} where $(x,\dot{x})$ are the robot position and velocity. The control input $\tau\in\mathbb{R}^2$ represents the torque applied to the robot. 
The initial position and goal position of the mobile robot are denoted as $x_0\in\mathscr{N}^2$ and $x_d\in\mathscr{N}^2$, respectively.
%The initial position of the mobile robot is denoted as $x_0\in\mathscr{N}^2$  and the goal position as $x_d\in\mathscr{N}^2$.
\subsection{Conformal Map}

Let $U$ be a connected open set, and $z=x+iy$ be a complex variable. A complex-valued function $f(z):U\subset\mathbb{C}\to\mathbb{C}$ is holomorphic on $U$ if it is complex differentiable at every point of $U$. 
%Let $z=x+iy$ be a complex variable, $U$ be a connected open set on $\mathbb{C}$, a complex-valued function $f:U\subset\mathbb{C}\to\mathbb{C}$ is holomorphic on $U$ if it is complex differentiable at every point of $U$. 

\textit{Definition 3 \cite{20}:} Let $f(x,y)=(u(x,y),v(x,y)):U\subset\mathbb{R}^2\to\mathbb{R}^2$ be of class $C^1$ with non-vanishing Jacobian, then $f$ is a conformal map if and only if $f(z)=u(x,y)+iv(x,y):U\subset\mathbb{C}\to\mathbb{C}$, as a function of $z$, is holomorphic.

\section{Conformal Navigation Transformation} %CONFORMAL NAVIGATION TRANSFORMATION

In this section, we define the conformal navigation transformation and investigate the existence and uniqueness of the proposed transformation.
%This section gives the definition of the conformal navigation transformation and studies the existence and uniqueness of the conformal navigation transformation.

Since the navigation function is constructed in a geometrically simple sphere world, the conformal navigation transformation defined in this paper provides an efficient way to implement the navigation function in a geometrically complex workspace. Formally, the conformal navigation transformation is defined as follows.

\textit{Definition 4:} A conformal navigation transformation is a map $T:\mathscr{N}^2\to\mathscr{M}^2$ which:
\begin{enumerate}
	\item maps the interior of the workspace conformally to the interior of a $2$-dimensional sphere world.
	\item maps the external boundary of the workspace homeomorphically to the unit circle.
	\item maps the boundaries of $M$ internal obstacles  homeomorphically to $M$ small disjoint circles inside the unit circle.
\end{enumerate}

This definition completely captures the topology of $2$-dimensional workspaces. According to the classification theorem of compact surfaces with boundary \cite{gallier2013}, two compact surfaces with boundary are topologically equivalent if and only if they agree in character of orientability, number of boundaries, and Euler characteristic. Obviously, $\mathscr{N}^2$ and $\mathscr{M}^2$ have the same orientability and number of boundaries. Moreover, the Euler characteristic of a workspace $\mathscr{N}^2$ with $M$ internal obstacles is $1-M$, which is the same as the Euler characteristic of the sphere world $\mathscr{M}^2$. Therefore, $\mathscr{N}^2$ and $\mathscr{M}^2$ are topologically equivalent.

In addition to being consistent with the topological relationship between the workspace and the sphere world, the conformal navigation transformation has several properties that enable it to be applied to robot navigation problems in complex workspaces.

The first property is that it exists in all workspaces in $\mathbb{R}^2$, implying that conformal navigation transformations can be constructed for 2D environments with arbitrarily shaped obstacles.
%The first property of the conformal navigation transformation is that it exists in all workspaces in $\mathbb{R}^2$. This implies that conformal navigation transformations can be constructed for 2D environments with arbitrarily shaped obstacles, and solutions to the path and motion planning problems in geometrically complex $2D$ environments can be obtained by applying the control laws proposed in Section \uppercase\expandafter{\romannumeral4}. 
The following results describe the first property in detail.

\textit{Lemma 1 \cite{21}:} Let $\Omega$ be a domain in Riemann sphere $\bar{\mathbb{C}}$ and let $B_{nc}(\Omega)$ be the collection of all boundary components of $\Omega$ which are not circles or points. If the closure of $B_{nc}(\Omega)$ in the space $B(\omega)$ of all boundary components of $\omega$ is (at most) countable, then $\Omega$ is conformally equivalent to a domain in $\bar{\mathbb{C}}$ whose boundary components are all circles or points.

The $2$-dimensional workspace is a closed set in Euclidean space containing all its $M+1$ boundaries, which is different from the multiply-connected open set in the Riemann sphere $\bar{\mathbb{C}}$. The following proposition generalizes Lemma $1$ to $\mathscr{N}^2$.

%Lemma 1 is applicable to the multi-connected domain in the Riemann sphere $\bar{\mathbb{C}}$, which is different from the 2-dimensional navigation space in the following two points. Firstly, the Riemann sphere is a model of the extended complex plane, which consists of the complex plane plus one point at infinity, and the $2$-dimensional navigation space is a subset of the Euclidean space. Secondly, the multi-connected domain in the Riemann sphere $\bar{\mathbb{C}}$ is an open set, while the 2-dimensional navigation space is a closed set containing all its $M+1$ boundaries.

\textit{Proposition 1 (Existence):} If $\mathscr{N}^2$ is a $2$-dimensional workspace with $M$ internal obstacles, then there exists a $2$-dimensional sphere world $\mathscr{M}^2$ with $M$ internal obstacles and a conformal navigation transformation $T$, such that $T(\mathscr{N}^2)=\mathscr{M}^2$. 
%If $\mathscr{W}^2$ is a $2$-dimensional workspace with $M$ internal obstacles, then there exists a $2$-dimensional sphere world $\mathscr{M}^2$ with $M$ internal obstacles and a conformal navigation transformation $T$, such that $T(\mathscr{W}^2)=\mathscr{M}^2$.

\begin{proof}  
Let $F:\Omega\to\Psi$ be a conformal map from a domain $\Omega$ in $\bar{\mathbb{C}}$ whose boundaries $\partial\Omega_i,\ i\in\{0,...,M\}$, are all Jordan curves onto a domain $\Psi$ in $\bar{\mathbb{C}}$ whose boundaries $\partial\Psi_i$ are all circles. According to Lemma 1, $F$ that satisfies the above requirements always exists. Let $G$ be the stereographic projection from $\bar{\mathbb{C}}\setminus\{z_0\}$ to 
the plane $\mathbb{R}^2$, and $H$ be the stereographic projection from $\bar{\mathbb{C}}\setminus\{F(z_0)\}$ to the plane $\mathbb{R}^2$.
Without loss of generality, it is assumed that the projection points $z_0$ and $F(z_0)$ are located in the interior of $\partial\Omega_0$ and $\partial\Psi_0$, respectively. For the stereographic projection, we know that it is both a diffeomorphism and a conformal map. Hence, the image $G(\Omega)$ is a bounded open set in $\mathbb{R}^2$ enclosed by a large Jordan curve  and $M$ small Jordan curves, meaning that $G(\Omega)$ is exactly the interior of a $2$-dimensional 
workspace in $\mathbb{R}^2$. Moreover, the image $H(\Psi)$ after normalization is a bounded open set in $\mathbb{R}^2$ enclosed by the unit circle $S^1$ and $M$ small circles in the interior of the unit circle, that is, $H(\Psi)$ is exactly the interior of a $2$-dimensional sphere world in $\mathbb{R}^2$. Since both $G$ and $H$ are stereographic projection and $F$ is a conformal map, the composition map $H\circ F\circ G^{-1}$ is a conformal map from $G(\Omega)$ to $H(\Psi)$. Defining $T$ as $T\triangleq H\circ F\circ G^{-1}$, then there always exists a conformal map $T$ which maps $\mathring{\mathscr{N}^2}$ onto $\mathring{\mathscr{M}^2}$. 
%$\mathring{\mathscr{N}^2}$ $\mathring{\mathscr{M}^2}$

%After obtaining the conformal map $T$ from $\mathring{\mathscr{N}^2}$ to $\mathring{\mathscr{M}^2}$, 
Let us next focus on the map from $\partial\mathscr{O}_i$ to $\partial\mathscr{\tilde{O}}_i$. For the workspace with $M$ obstacles, we know that it can easily be reduced to the simply connected case by making suitable cuts \cite{pommerenke1992}. %For the workspace with $M$ obstacles, it has the same boundary behavior as the workspace without obstacles. To see this, choose a ring domain in the neighborhood of a boundary component in $\mathscr{N}^2$, after an additional map with a suitable logarithm we have a simply-connected domain. 
So we get a reduction of the boundary behavior to the case of a workspace without obstacles. Without loss of generality, let $\partial\mathscr{O}_0$ be the only boundary of the ${\mathscr{N}^2}$ without obstacles. Since $\partial\mathscr{O}_0$ is a Jordan curve, it can be represented by the following parameterization: $\partial\mathscr{O}_0=T^{-1}(e^{it}), i\in [0,2\pi]$, and $T^{-1}$ is a continuous bijective map from the unit circle $S^1$ onto $\partial\mathscr{O}_0$. Hence, $T$ is a homeomorphism from $\partial\mathscr{O}_0$ to $S^1$ since a continuous bijective map on a compact space is a homeomorphism onto its image. Applying the boundary properties of a workspace without obstacles to a workspace with $M$ obstacles, the conclusion can be drawn that $T$ maps $\partial\mathscr{O}_0$ homeomorphically to $S^1$ and $\partial\mathscr{O}_i$ homeomorphically to $M$ small disjoint circles inside $S^1$. Finally, since $T$ satisfies all conditions of Definition 4, there always exists a conformal navigation transformation that maps ${\mathscr{N}^2}$ onto $\mathscr{M}^2$.
\end{proof}
%$\hfill\qedsymbol$

After concluding that conformal navigation transformations are ubiquitous in $2$-dimensional workspaces, it is natural to wonder how many conformal navigation transformations exist in a given $2$-dimensional workspace. Proposition 2 answers this question and leads to the second property of the conformal navigational transformation: uniqueness.

%\textit{Proposition 2:} For a 2-dimensional navigation space ${\mathscr{N}^2}$ with $M$ internal obstacles, there exists a  $2$-dimensional sphere world $\mathscr{M}^2$ with $M$ internal obstacles and a  conformal navigation transformation $T:\mathscr{N}^2\to\mathscr{M}^2$. Both $\mathscr{M}^2$ and $T$ are uniquely determined by ${\mathscr{N}^2}$ when the normalization condition is imposed:

%The images of  the prescribed three points on the outer boundary $\partial\mathscr{O}_0$ are fixed  on the unit circle.

\textit{Proposition 2 (Uniqueness):} The conformal navigation transformation $T:\mathscr{N}^2\to\mathscr{M}^2$ and $\mathscr{M}^2$ are uniquely determined by ${\mathscr{N}^2}$ if the images of the prescribed three points on the external boundary $\partial\mathscr{O}_0$ are fixed on the unit circle.

\begin{proof}
See Appendix A.
\end{proof}

%\textit{Proof:} See Appendix A.$\hfill\qedsymbol$

The uniqueness of conformal navigation transformations has two advantages when applied to robot navigation problems. Firstly, no further parameter tuning is required after the construction of the conformal navigation transformation, that is, the resulting solution is correct-by-construction. Secondly, the conformal navigation transformation is easy to determine. In fact, it can be uniquely determined by fixing only three points on the unit circle.

\section{Controller %Controller 
Design Using the Conformal Navigation Transformation}% CONTROLER DESIGN USING THE CONFORMAL NAVIGATION TRANSFORMATION
%UTILIZING
%In the previous section we have shown that the conformal navigation transformation is uniquely existing in the 2D environment. Moreover, we have known some of its advantages in the design of controller for the robot navigation problem. 
In this section, the third property, namely the invariance of the navigation function properties, is given, then two controllers based on the conformal navigation transformation are presented and analyzed to illustrate its application to the robot navigation problem in complex 2D environments.
%In this section, the third property of the conformal navigation transformation is given, and then two controlers based on the conformal navigation transformation are presented and analyzed to illustrate its application to the robot navigation problem in complex $2D$ environments.
%\subsection{Navigation properties invariance}
%The third property of conformal navigation transformations is necessary for the design of control methods and is described by Proposition 3 as follows.

The third property is important for extending the construction of navigation functions on sphere worlds to any complex 2D environment, as described in Proposition 3.
%The third property that is important for applying conformal navigation transformations to design control laws for navigation problems in complex 2D environments, as described in Proposition 3.

\textit{Proposition 3 (invariance):} Let $\mathscr{N}^2$ be a workspace with smooth boundaries, $T:\mathscr{N}^2\to\mathscr{M}^2$ be the unique conformal navigation transformation determined by $\mathscr{N}^2$, and $\phi:\mathscr{M}^2\to[0,1]$ be a navigation function on $\mathscr{M}^2$. Then $\tilde{\phi}=\phi\circ T$ is a navigation function on $\mathscr{N}^2$.

\begin{proof}%\textit{Proof:} 
Since every $\partial\mathscr{O}_i$ is a smooth Jordan curve, according to the Kellogg-Warschawski theorem \cite{pommerenke1992}, $T$ is a smooth map on $\mathscr{N}^2$. Then, $\tilde{\phi}$ is a smooth function because it is a composition of smooth function $\phi$ and smooth map $T$. 

According to Lemma1, the composition map $T\triangleq H\circ F\circ G^{-1}$ is a conformal equivalence from $\mathring{\mathscr{N}^2}$ onto $\mathring{\mathscr{M}^2}$. Therefore, $T$ is a diffeomorphism from $\mathring{\mathscr{N}^2}$ onto $\mathring{\mathscr{M}^2}$ and the Jacobian of $T$ is non-singular in $\mathring{\mathscr{N}^2}$. It is sufficient to study the influence of $T$ on the critical points of the navigation function on $\mathring{\mathscr{N}^2}$ because the critical points of the navigation function $\phi$ are all in $\mathring{\mathscr{N}^2}$. Applying the chain rule yields $\nabla\tilde{\phi}=[J_T]^T\nabla(\phi\circ T)$. As $T$ is non-singular in $\mathring{\mathscr{N}^2}$, $T$ is a bijection from the set of critical points of $\tilde{\phi}$, $\mathscr{C}_{\tilde{\phi}}$, to the set of critical points of $\phi$, $\mathscr{C}_{\phi}$. Furthermore, the Hessian matrix of $\tilde{\phi}$ at the critical points satisfies $H(\tilde{\phi})=[J_T]^T H(\phi\circ T)[J_T]$. Since $T$ is non-singular in $\mathring{\mathscr{N}^2}$, $H(\tilde{\phi})$ is non-singular at the critical points. Hence, $\tilde{\phi}$ is a Morse function.

Next, consider the minimum of $T$ in $\mathscr{N}^2$. At the critical points, we have that: $H(\tilde{\phi})=[J_T]^T H(\phi\circ T)[J_T]$. Since $T$ is a bijection between $\mathscr{C}_{\tilde{\phi}}$ and $\mathscr{C}_{\phi}$, the Morse index of $\tilde{\phi}$ matches the Morse index of $\phi$ at each of the critical points. Furthermore, $\phi$ has a unique minimum at $x_d$. Hence, $\tilde{\phi}$ has exactly one minimum at $p_d=T(x_d)$.

Last consider the image of $\partial\mathscr{O}_i$ under $\tilde{\phi}$. Since $T$ is a diffeomorphism on $\mathring{\mathscr{N}^2}$ and a homeomorphism on each smooth boundary $\partial\mathscr{O}_i$, the image of $\partial\mathscr{O}_i$ under $T$ is exactly $\partial\mathscr{\tilde{O}}_i$. Moreover, $\phi$ is uniformly maximal on $\partial\mathscr{O}_i$. Hence, $\tilde{\phi}$ is admissible on $\mathscr{N}^2$.
\end{proof}
%$\hfill\qedsymbol$

\textit{Remark 1:} For Proposition 3, the smoothness of the boundaries of $\mathscr{N}^2$ is necessary because the navigation function $\phi$ is smooth on the boundaries of $\mathscr{N}^2$. However, this does not prevent the application of the conformal navigation transformation to $\mathscr{N}^2$ with non-smooth boundaries. In fact, the pullback of a trivial solution on the model space by $T$ is a solution on the original space \cite{19}.

Proposition 3 shows that the conformal navigation transformation between $\mathscr{M}^2$ and $\mathscr{N}^2$ with smooth boundaries does not change the navigation function properties. %This property is important for the application of conformal navigation transformation to the design of control laws for navigation problems in complex two-dimensional environments.
This property provides an approach to constructing navigation functions in complex 2D environments, as shown below.

\textit{Proposition 4:} The mobile robot system \eqref{kinematic} under the feedback control law \begin{equation}\begin{split}\label{}u(x)&=-K(J_T(x))^T\nabla_{T(x)}\phi_{kr}(T(x))\\&=-K\lVert J_T(x)\rVert(J_T(x))^{-1}\nabla_{T(x)}\phi_{kr}(T(x)),\end{split}\label{Proposition 4 control law }\end{equation} where $K$ is a positive gain, is globally asymptotically stable at the goal point $x_d\in\mathring{\mathscr{N}^2}$ from almost every initial point in $\mathring{\mathscr{N}^2}$ with smooth boundaries.

\begin{proof} %\textit{Proof:} 
Since $T$ is a conformal map on $\mathring{\mathscr{N}^2}$, according to the Cauchy-Riemann equation, the Jacobian $J_T(x)=\begin{bmatrix}{u_x}_1&-{v_x}_1\\{v_x}_1&{u_x}_1\end{bmatrix}$ at $x=(x_1,x_2)\in\mathring{\mathscr{N}^2}$. Moreover, the Jacobian $J_T(x)={({{u_x}_1}^2+{{v_x}_1}^2)}^{\frac{1}{2}}R={\lVert J_T(x)\rVert}^{\frac{1}{2}}R$, where $R$ is a rotation matrix. Hence, for $T$ on $\mathring{\mathscr{N}^2}$, we have that: $(J_T(x))^T=\lVert J_T(x)\rVert(J_T(x))^{-1}$. 

Choosing $V(x)=\phi_{kr}(T(x))$ as a candidate Lyapunov function. Differentiating $V$ along the system's trajectories yields: $\dot{V}(x)=-K\lVert J_T(x)\lVert\lVert\nabla_{T(x)}\phi_{kr}(T(x))\lVert^2\leqslant0.$
Since $J_T(x)$ is non-singular, the set for $\dot{V}(x)=0$ consists only of the critical points of $\phi_{kr}(T(x))$. The critical points of $\phi_{kr}(T(x))$ include the unique minimum point $x_d$ and the isolated saddle points. Lasalle's Invariance Theorem dictates that the closed loop system will converge either to the goal or the saddle points. Since the saddle point has a stable manifold of 1-dimension in $\mathring{\mathscr{N}^2}$, the isolated saddle points have attractive basins of zero measure. Hence $\dot{V}(x)<0$ holds for almost everywhere, which implies global asymptotic stability for the system from almost all initial points.
\end{proof}
%$\hfill\qedsymbol$

\textit{Remark 2:} Let $r_d=\lVert q-q_d\rVert^2$. Taking the gradient of $\phi_{kr}$ at $q\in\mathscr{M}^2$, we have that \begin{equation}\begin{split}\nabla\phi_{kr}&=2\beta\left(\lVert q-q_d\rVert^{2k}+\beta\right)^{-(\frac{k+1}{k})}\Big((q-q_d)\\& \quad -\frac{\lVert q-q_d\rVert^2}{k}\sum_{i=0}^M\frac{1}{|\beta_i|}(q-q_i)\Big)\\&=2\beta\left( r_d^k+\beta\right)^{-(\frac{k+1}{k})}\Big((q-q_d)\\& \quad -\frac{r_d}{k}\sum_{i=0}^M\frac{1}{|\beta_i|}(q-q_i)\Big).\end{split}\end{equation}
Since $\beta\left( r_d^k+\beta\right)^{-(\frac{k+1}{k})}$ is a positive scalar, the direction of $\nabla\phi_{kr}$ is the same as the direction of the weighted average of $q-q_d$ and $q-q_i$, and the critical points of $\nabla\phi_{kr}$ are the points where the weighted average is a zero vector. Hence, any point in the set \begin{equation}\mathscr{S}=\{q\in\mathring{\mathscr{M}^2}:(q-q_d)-\frac{r_d}{k}\sum_{i=0}^M\frac{1}{|\beta_i|}(q-q_i)=0,q\neq q_d\}\nonumber\end{equation} is a saddle point of $\nabla\phi_{kr}$, and any point $x\in\mathring{\mathscr{N}^2}$ mapped in the set $\mathscr{S}$ is a saddle point of $u(x)$.

The velocity of the robot at each point $x$ in $\mathring{\mathscr{N}^2}$ is a rotation and scaling of the velocity of the corresponding point $T(x)$ in $\mathring{\mathscr{M}^2}$. To see this, observe that $(J_T(x))^T$ is a positive scalar $\lVert J_T(x)\rVert^{\frac{1}{2}}$ times a rotation matrix $R^T$. This implies that the conformal navigation transformation preserves both angle and local shape. Therefore, the conformal navigation transformation can completely eliminate the angular distortion when mapping from a geometrically complex workspace to a sphere world, which is its fourth property.
%To see this, observe that for the conformal navigation transformation $T$, the transpose of its Jacobian, $(J_T(x))^T$, is a positive scalar $\lVert J_T(x)\rVert}^{\frac{1}{2}}$ times a rotation matrix $R^T$. Furthermore, the Jacobian of $T$ is also a composition of a rotation and a scaling, which means that the conformal navigation transformation preserves both angular and local shape.

For the mobile robot with mass $m$ and second-order integrator dynamics, a minor modification can be used, as stated in the following proposition.

\textit{Proposition 5:} The mobile robot system \eqref{dynamic} under the feedback control law \begin{equation}\begin{split}\label{}u(x)&=-K(J_T(x))^T\nabla_{T(x)}\phi_{kr}(T(x))+d(x,\dot x)\\&=-K\lVert J_T(x)\rVert(J_T(x))^{-1}\nabla_{T(x)}\phi_{kr}(T(x))+d(x,\dot x),\end{split}\label{Proposition 5 control law }\end{equation} where $K$ is a positive gain and $d(x,\dot x)$ is a dissipative field, i.e., $\dot{x} ^T d(x,\dot{x})<0$ for $\dot{x} \neq 0 $, is globally asymptotically stable at the goal point $x_d\in\mathring{\mathscr{N}^2}$ from almost every initial point with zero initial velocity in $\mathring{\mathscr{N}^2}$ with smooth boundaries.

\begin{proof} %\textit{Proof:} 
Since $\phi_{kr}(T(x))$ is a navigation function on $\mathscr{N}^2$ and $\dot{x}_0(t)=0$ at $t=0$, the trajectory of the mobile robot system \eqref{dynamic} under the feedback control law \eqref{Proposition 5 control law } stay inside $\mathscr{N}^2$ \cite{Rimon1992}. Consider the candidate Lyapunov function:\begin{equation}V(x)=K\phi_{kr}(T(x))+\frac{1}{2}m\dot{x}^T \dot{x}.\nonumber \end{equation}
Its time derivative is provided as: \begin{equation}\dot{V}(x)=K[\nabla_{T(x)}\phi_{kr}(T(x))]^T J_T(x)\dot{x}+m\ddot{x}^T\dot{x} .\nonumber \end{equation} Following the trajectory of the closed-loop system, we get $\dot{V}(x)=[d(x,\dot{x})]^T \dot{x}\leqslant 0$. Noting that the set where $\dot{V}(x)=[d(x,\dot{x})]^T \dot{x}=0$ is equal to the set where $\dot{V}(x)=0$. Hence, the largest invariant set consists only of the goal point $x_d$ and the saddle points. Since only a set of measure zero of initial points is attracted to the
saddles, then according to the Lasalle's Invariance Theorem, $\dot{V}(x)<0$ holds for almost everywhere. The control law \eqref{Proposition 5 control law } is thus globally asymptotically stable at $x_d$ from almost every point in  $\mathring{\mathscr{N}^2}$.
\end{proof}
%$\hfill\qedsymbol$

\textit{Remark 3:} The dissipation term in (8) can also be chosen as $d(T(x),\dot{T}(x))$, which does not change the system's stability at the equilibrium point but introduces complex tensor calculations when proving the stability. In fact, $d$ can be an arbitrary dissipative vector field, and the critical qualitative behavior of the robotic system (5) is essentially the same as that of much simpler gradient systems \cite{22}.

The above results provide an approach to solving the robot navigation problem in complex workspaces: the navigation function $\phi$ in $\mathscr{N}^2$ is pulled back to $\mathscr{M}^2$ by a conformal navigation transformation $T$, and its gradient is used  to construct the control law to obtain a solution to the motion planning problem. 
Since $T$ is a diffeomorphism on $\mathring{\mathscr{N}^2}$, there is an alternative way of solving the above problem: pulling back the solution determined by a vector field on $\mathscr{M}^2$ to $\mathscr{N}^2$ by $J_T^{-1}$. It is worth mentioning that the special property of $T$, i.e., $[J_T]^T=\lVert J_T\rVert [J_T]^{-1}$, allows establishing a connection between the above two methods.%, which will be explained in detail in a paper currently under preparation.

\section{Construction of the Conformal Navigation Transformation}

In this section, we present a construction method of the conformal navigation transformation without tuning parameters, which can be operated directly in the 2D environment with non-star-shaped obstacles. We first construct $T$ in workspaces without obstacles and then use them successively to obtain $T$ in workspaces with $M$ internal obstacles. The machinery of complex variables facilitates the construction of conformal navigation transformations in $\mathscr{N}^2$, which will be used in this section. In fact, $T(x)$ on $\mathscr{N}^2$ and $T(z)$ on the complex plane satisfy the relation: $T(x)=
[\Re(T(z)),\Im(T(z))]$.%$T(x)=(\Re(T(z)),\Im(T(z)))$

\subsection{Conformal navigation transformations for simply connected workspaces}

Let $U^+$ and $U^-$ be a bounded and an unbounded simply connected workspace in the extended complex plane $\mathbb{C}\cup\{\infty\}$, respectively. 
For $U^+$, assume that $z_c$ is a given point in $U^+$. For $U^-$, assume that $z_c$ is a given point in the complement of $U^-$ and $\infty\in U^-$. The boundary $\Gamma$ is a closed smooth Jordan curve which is parametrized by a $2\pi$-periodic twice continuously differentiable complex
function $\gamma (s)$ with $\gamma'(s)\neq 0$ for $s\in [0,2\pi]$. The orientation of $\Gamma$ is counterclockwise for $U^+$ and clockwise for $U^-$. 

According to Definition 3 and Definition 4, the conformal navigational transformation $T$ in the extended complex plane is a holomorphic function $T(z)$ with a non-zero derivative and has a continuous extension to the boundary $\Gamma$.%According to Definition 3 and Definition 4, the conformal navigational transformation $T(z)$ in the extended complex plane is a holomorphic function with a non-zero derivative and has a continuous extension to the boundary $\Gamma$. 
Then, from the Cauchy integral formula we have that:

\begin{equation} T(z)=\frac{1}{2\pi i} \int_{\Gamma}\frac{T(\gamma)}{\gamma-z}d\gamma, \ {z\in U^+},\label{Cauchy integral formula 1}\end{equation}
and 
\begin{equation} T(z)=T(\infty)-\frac{1}{2\pi i} \int_{\Gamma}\frac{T(\gamma)}{\gamma-z}d\gamma, \ {z\in U^-}.\label{Cauchy integral formula 2}\end{equation}
%\begin{equation} T(z)=\left\{\begin{array}{rcl}\frac{1}{2\pi i} \int_{\partial\mathscr{O}_0}\frac{T(\gamma)}{\gamma-z}d\gamma & & {z\in U^+}\\T(\infty)-\frac{1}{2\pi i}\int_{\partial\mathscr{O}_0}\frac{T(\gamma)}{\gamma-z} d\gamma & & {z\in U^-} \end{array} \ringt.\end{equation}
%\begin{eqnarray} T(z)=\begin{cases}\frac{1}{2\pi i} \int_{\partial\mathscr{O}_0}\frac{T(\gamma)}{\gamma-z}d\gamma, & {z\in U^+},\\ T(\infty)-\frac{1}{2\pi i}\int_{\partial\mathscr{O}_0}\frac{T(\gamma)}{\gamma-z} d\gamma, & {z\in U^-}. \end{cases}\end{equation}

Equations \eqref{Cauchy integral formula 1} and \eqref{Cauchy integral formula 2} imply that the conformal navigation transformations on $U^+$ and $U^-$ are completely determined by their values on boundary $\Gamma$, which is the fifth property of $T$. This property reduces the construction problem of $T$ on the entire workspace to the construction problem of $T$ on the boundary. Therefore, we focus on the solution of $T(\gamma (s))$.

For a bounded simply connected workspace $U^+$, its boundary $\Gamma$ is mapped onto the unit circle $\partial D$ and $T(\gamma (s))$ is given by:
\begin{equation} T(\gamma (s))=e^{i\theta (s)}, \ s\in [0,2\pi].\label{boundary value}\end{equation} 
The proof of Proposition 2 shows that the condition for $T(\gamma (s))$ to take the unique value is equivalent to:\begin{equation} T(z_c)=0, \ T'(z_c)>0.\label{unique value}\end{equation} 
Using the properties of the complex exponential function $e^{f(z)}$ and \eqref{unique value}, $T(z)$ is constructed as follows:{\begin{equation} T(z)=T'(z_c)(z-z_c)e^{(z-z_c)f(z)},\label{T(z)}\end{equation}}where $f(z)$ is a holomorphic function on $U^+$. Taking the logarithm of both sides of \eqref{T(z)} and substituting \eqref{boundary value}, we have that: \begin{equation}\begin{split} (\gamma (s)-z_c)f(\gamma (s))=&-\log(|\gamma (s)-z_c|)-\log({T}'(z_c))\\&+i(\theta (s)-\arg(\gamma (s)-z_c)).\end{split}\label{integral}\end{equation}

Equation \eqref{integral} can be written as: \begin{equation}G(s)f(\gamma (s))=\mu(s)+c(s)+i\upsilon(s), \ s\in [0,2\pi],\label{integral simple}\end{equation} where $G(s)=\gamma (s)-z_c$, $\mu(s)=-\log|\gamma (s)-z_c|$, $c(s)=-\log\big(T'(z_c)\big)$, $\upsilon(s)=\theta (s)-\arg\big(\gamma (s)-z_c\big)$. Note that $\mu(s)$ is a given real-valued function and $c(s)$ is a constant. Although $f(\gamma (s))$, $c(s)$, and $\upsilon(s)$ are all unknown, \eqref{integral simple} is still solvable since $\mu(s)$ is a known real-valued function and $f(z)$ is a holomorphic function on $U^+$. The following lemma %Lemma 2% 
gives the above conclusion in more detail. The various variables mentioned in the sequel are defined in Appendix B.

\begin{figure*}[!t]
   \centering
	\includegraphics[width=17cm]{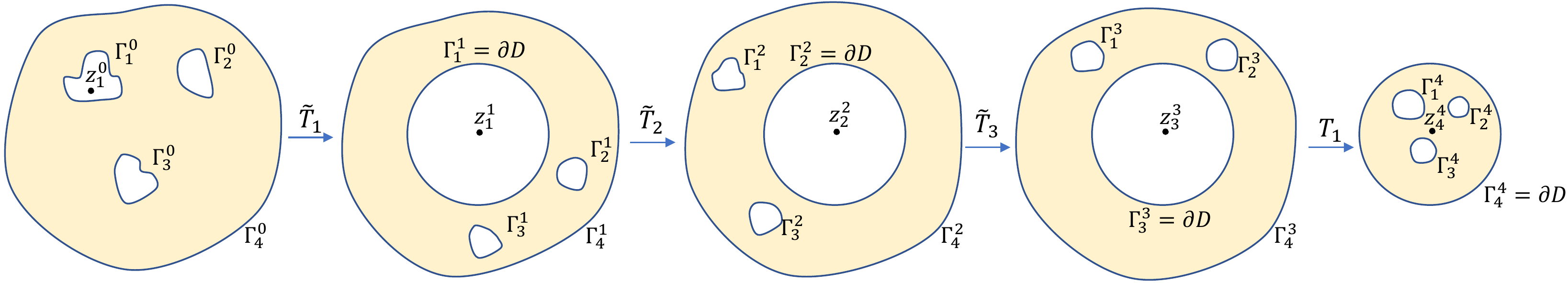}
	\caption{First iteration in a workspace with 3 internal obstacles.}
	\label{Fig_1}
\end{figure*}

{\textit{Lemma 2:}} For the given real-valued function $\mu$, there exists a unique constant $c$ and a unique function $\upsilon$, such that \eqref{integral simple} are boundary values of a holomorphic function $f$ in $U^+$. The function $\upsilon$ is the unique solution of the equation \begin{equation} \upsilon-\mathbf R\upsilon=-\mathbf H\mu\label{integral equation}\end{equation} and $c$ is given by \begin{equation} c=[\mathbf H \upsilon-(\mathbf I-\mathbf R)\mu]/2.\label{c}\end{equation}
%For the given real-valued function $\mu(s)$, there exists a unique constant $c$ and a unique holomorphic function $f(z)$ in $U^+$, such that \eqref{integral simple} is solvable. The function $\upsilon(s)$ is the unique solution of the equation \begin{equation} \mu-\mathbf R\mu=-\mathbf H \upsilon\label{integral equation}\end{equation} and $c$ is given by \begin{equation} c=[\mathbf H \mu-(\mathbf I-\mathbf R)\upsilon]/2.\label{c}\end{equation}
%For the given real-valued function $\mu$, there exists a unique constant $c$ and a unique function $\upsilon$, such that \eqref{integral simple} are boundary values of an holomorphic function $f$ in $U^+$. The function $\upsilon$ is the unique solution of the equation \begin{equation} \mu-\mathbf R\mu=-\mathbf H \upsilon\label{integral equation}\end{equation} and $c$ is given by \begin{equation} c=[\mathbf H \mu-(\mathbf I-\mathbf R)\upsilon]/2.\label{c}\end{equation}

\begin{proof}%\textit{Proof:} 
See Appendix B.
\end{proof}
%$\hfill\qedsymbol$

Since \eqref{integral simple} is uniquely solvable, the conformal navigation transformation $T(z)$ on $U^+$ can be solved by \eqref{T(z)}, as described in Proposition 6.

\textit{Proposition 6:} For any bounded simply connected workspace $U^+$, and any twice continuously differentiable parametric boundary $\gamma (s)$ with $\gamma'(s)\neq 0$, the holomorphic function
\begin{equation} T(z)=e^{-c}(z-z_c)e^{(z-z_c)f(z)}\end{equation} is the conformal navigation transformation on $U^+$.
%\begin{equation}\begin{split} T(z)&=e^{-c}(z-z_c)e^{(z-z_c)f(z)}\\&=\frac{1}{2\pi i} \int_{\Gamma}\frac{e^{i\Big(\upsilon (s)+\arg\big(\gamma (s)-z_c\big)\Big)}}{\gamma-z}d\gamma.\end{split}\end{equation} is the conformal navigation transformation on $U^+$.

\begin{proof}%\textit{Proof:} 
According to Lemma 2, the unknown constant $c=-\log\big(T'(z_c)\big)$ is given by \eqref{c}. Thus, we have that $T'(z_c)=e^{-c}$. Then, in view of \eqref{T(z)}, we obtain that: $T(z)=e^{-c}(z-z_c)e^{(z-z_c)f(z)}$. Differentiating $T(z)$ with $z$, we get that: \begin{equation}\begin{split} T'(z)=&e^{-c}(e^{(z-z_c)f(z)}+\\&(z-z_c)e^{(z-z_c)f(z)}\big((z-z_c)f'(z)+f(z)\big)\end{split}\end{equation} and $T'(z)\neq 0$ in the interior of $U^+$. Since $T(\gamma (s))=e^{i\theta (s)}$, the function $T(\gamma (s))$ is a homeomorphism on $\Gamma$. Hence, $T(z)$ is the conformal navigation transformation on $U^+$.
\end{proof}
%$\hfill\qedsymbol$

%Lemma 2 shows that the unknown function $\upsilon (s)=\theta (s)-\arg\big(\gamma (s)-z_c\big)$ is the unique solution of equation (16) where the function $\mu(s)=-log|\gamma (s)-z_c|$ is known. Thus the function $\theta (s)$ is given by $\theta (s)=\upsilon (s)+\arg\big(\gamma (s)-z_c\big)$. Since $T(z)$ is a holomorphic function and $\Gamma$ is a closed smooth
%ordan curve, then according to Cauchy integral formula, the conformal navigtion function $T(z)$ can be written as  \begin{equation}T(z)=\frac{1}{2\pi i} \int_{\Gamma}\frac{e^{i\Big(\upsilon (s)-\arg\big(\gamma (s)-z_c\big)\Big)}}{\gamma-z}d\gamma.\nonumber\end{equation}$\hfill\qedsymbol$

Unlike $T(z)$ on $U^+$, the conformal navigation transformation on the unbounded simply connected workspace $U^-$, denoted as $\tilde{T}(z)$, maps $U^-$ to the exterior of the unit circle. The boundary values of $\tilde{T}(z)$ satisfy
\begin{equation} \tilde{T}(\gamma (s))=e^{-i\theta (s)}, \ s\in [0,2\pi].\label{boundary values 2}\end{equation} 
With the condition
\begin{equation} {T}(\infty)=\infty, \ \tilde{T}'(\infty)>0,\end{equation} the function $\tilde{T}(z)$ is unique in $U^-$.

%\addtolength{\topmargin}{0.06in}

Similar to $T(z)$ on $U^+$, the function $\tilde{T}(z)$ on $U^-$ is constructed as
\begin{equation} \tilde{T}(z)=\tilde{T}'(\infty)(z-z_c)e^{-\tilde{f}(z)},\end{equation} 
where $\tilde{f}(z)$ is a holomorphic function on $U^-$ and $\tilde{f}(\infty)=0$.
From \eqref{boundary values 2}, we get that:
\begin{equation}\begin{split} \tilde{f}\big(\gamma (s)\big)=& \log\big(|\gamma (s)-z_c|\big)+\log\big(\log\big(\tilde{T}'(\infty)\big)\\&+i\Big(\theta (s)+\arg\big(\gamma (s)-z_c\big)\Big).\end{split}\end{equation} 
Let $\tilde{\mu}(s)=\log|\gamma (s)-z_c|$, $\tilde{c}(s)=\log\big(\tilde{T}'(\infty)\big)$, and $\tilde{\upsilon}(s)=\theta (s)+\arg\big(\gamma (s)-z_c\big)$, then we obtain that:
\begin{equation}\tilde{f}(\gamma (s))=\tilde{\mu}(s)+\tilde{c}(s)+i\tilde{\upsilon}(s), \ s\in [0,2\pi].\label{integral simple 2}\end{equation}
Equation \eqref{integral simple 2} has similar solvability as \eqref{integral simple}, which is described in Lemma 3.

\textit{Lemma 3:} For the given real-valued function $\tilde{\mu}$, there exists a unique constant $\tilde{c}$ and a unique function $\tilde{\upsilon}$ in $U^-$, such that \eqref{integral simple 2} are boundary values of a holomorphic function $\tilde{f}$ with $\tilde{f}(\infty)=0$. The function $\tilde{\upsilon}(s)$ is the unique solution of the equation \begin{equation} \tilde{\upsilon}-\mathbf R\tilde{\upsilon}=-\mathbf H\tilde{\mu} \end{equation} and $\tilde{c}$ is given by \begin{equation} \tilde{c}=[\mathbf H\tilde{\upsilon}-(\mathbf I-\mathbf R)\tilde{\mu}]/2.\end{equation}
%For the given real-valued function $\tilde{\mu}(s)$, there exists a unique constant $\tilde{c}$ and a unique holomorphic function $\tilde{f}(z)$ in $U^-$, such that \eqref{integral simple 2} are boundary values of $\tilde{f}(z)$ with $\tilde{f}(\infty)=0$. The function $\tilde{\upsilon}(s)$ is the unique solution of the equation \begin{equation} \tilde{\mu}-\mathbf R\tilde{\mu}=-\mathbf H \tilde{\upsilon}\end{equation} and $\tilde{c}$ is given by \begin{equation} \tilde{c}=[\mathbf H\tilde{\mu}-(\mathbf I-\mathbf R)\tilde{\upsilon}]/2.\end{equation}

\begin{proof}%\textit{Proof:} 
This lemma can be proved along the same lines as Lemma 2.
\end{proof}
%$\hfill\qedsymbol$

The solution of $\tilde{T}(z)$ on $U^-$ can be obtained by solving \eqref{integral simple 2}, as shown in Proposition 7 below.

\textit{Proposition 7:} For any unbounded simply connected workspace $U^-$, and any twice continuously differentiable parametric boundary $\gamma (s)$ with $\gamma'(s)\neq 0$, the holomorphic function
\begin{equation} \tilde{T}(z)=e^{\tilde{c}}(z-z_c)e^{-\tilde{f}(z)}\end{equation} is the conformal navigation transformation on $U^-$.
%\begin{equation}\begin{split} \tilde{T}(z)&=e^{\tilde{c}}(z-z_c)e^{-\tilde{f}(z)}\\&=\frac{1}{2\pi i} \int_{\Gamma}\frac{e^{i\Big(\arg\big(\gamma (s)-z_c\big)-\tilde{\upsilon} (s)\Big)}}{\gamma-z}d\gamma.\end{split}\end{equation} is the conformal navigation transformation on $U^-$.

\begin{proof}%\textit{Proof:} 
This proposition can be proved along the same lines as Proposition 6.
\end{proof}
%$\hfill\qedsymbol$

There is no unbounded simply connected workspace in the realistic 2D environment; however, $\tilde{T}(z)$ on $U^-$ is used in the following construction of the conformal navigation transformation in the multiply-connected workspace.

\subsection{Conformal navigation transformations for multiply-connected workspaces}

Let $N$ be a bounded workspace with $M$ internal obstacles in the extended complex plane $\mathbb{C}\cup\{\infty\}$. The boundary $\Gamma=\sqcup_{i=1}^{m+1}\Gamma_i$ is the disjoint union of  $M+1$ closed smooth Jordan curves and $\Gamma_{m+1}$ is the external boundary. We assume that $z_i$, $1\leqslant i\leqslant m$, is a given point inside $\Gamma_i$, $1\leqslant i\leqslant m$, and $z_{m+1}$ is a given point in the interior of $N$. The curve $\Gamma_i$ is parametrized by a $2\pi$-periodic twice continuously differentiable complex
function $\gamma_i(s)$ with $\gamma'_i(s)\neq 0$ for $s\in [0,2\pi]$. The orientation of $\Gamma$ is such that $N$ is always on the left of $\Gamma$.

The construction method of $T(z)$ on $N$ is an efficient and fast iterative method, which is inspired by the conventional Koebe's method on unbounded regions. The building blocks of the method are $T(z)$ on $U^+$ and $\tilde{T}(z)$ on $U^-$. Each iteration consists of the following $M+1$ steps, which are illustrated in Fig. 1 when $M$ is 3.

At the beginning, set $\Gamma_i^0=\Gamma_i$ and $z_i^0=z_i$ for $i=1,2,...,m+1.$ Then, the first step constructs the function $\tilde{T}_1$ that maps the exterior region of the curve $\Gamma_1^0$ onto the exterior of the unit circle and $z_1^0$ to the origin. Since the curves $\Gamma_i^0$, $i\neq 1$, are in the exterior region of $\Gamma_1^0$, the images of the curves $\Gamma_i^0, i\neq 1$, are smooth Jordan curves $\Gamma_i^1=\tilde{T}_1(\Gamma_i^0)$, $i\neq 1$,which can be computed using \eqref{Cauchy integral formula 1}. Repeat this, and at step $k\leqslant m$, construct the function $\tilde{T}_k$, that maps the exterior region of the curve $\Gamma_k^0$ onto the exterior region of the unit circle and $z_k^{k-1}$ to the origin. So far, every internal boundary has been mapped to the unit circle once. For the external boundary $\Gamma_{m+1}^m$, at the $(m+1)$-th step, construct the function $T_1$, that maps the interior region of the curve $\Gamma_{m+1}^m$ onto the open unit disk and $z_{m+1}^
m$ to the origin. Then, $\Gamma_{m+1}^{m+1}=T_1(\Gamma_{m+1}^m)$ is the unit circle and $\Gamma_i^{m+1}=T_1(\Gamma_i^m)$, $1\leqslant i\leqslant m$, are smooth Jordan curves, which can be computed using \eqref{Cauchy integral formula 2}.

The above process is repeated for each iteration, and as the number of iterations increases, each boundary becomes gradually more circular. Define $\bar{T}_1=T_1\circ\tilde{T}_m\circ\tilde{T}_{m-1}\circ\cdots\circ\tilde{T}_1$ as the function after the first iteration and define $\bar{T}_n=\hat{T}_n\circ\hat{T}_{n-1}\circ\cdots\circ\hat{T}_1$ as the approximate conformal navigation transformation after $n$ iterations. Then, the above assertion is equivalent to the fact that the function $\bar{T}_n\circ T^{-1}$ tends to the identity function as $n$ increases. The following proposition gives the quantitative convergence estimation. 

\textit{Proposition 8 \cite{WEGMANN2005}:} For the workspace $N$ with $M$ internal
obstacles, there exist constants $h>0$, $0<q<1$, such that for $n=1,2,\dots$ and for all $z\in N$,\begin{equation} \|\bar{T}_n\circ T^{-1}(z)-z\|\leqslant hq^{4n}.\end{equation}

\textit{Remark 4:} Since $0<p<1$, the iteration always converges. The convergence rate $q$ depends on the center and radius of each circular obstacle in the sphere world corresponding to $N$. Actually, the simulation experiments show that the rate of convergence is much better than $q$.

%\addtolength{\topmargin}{0.015in}

Since $T(z)$ is completely determined by its boundary values, the conformal navigation transformation has the advantage of being dependent on neither the initial nor the goal point, and does not need to be reconstructed for the changing goal point; it only needs to be constructed once.

\section{Simulation Results} %SIMULATION RESULTS

In order to validate the effectiveness of the proposed methodology, a set of simulation results was presented in this section. Simulations were carried out for both controllers \eqref{Proposition 4 control law } and \eqref{Proposition 5 control law }, and the same Koditschek-Rimon navigation function $\phi_{kr}$ was used. The parameter $k$ in $\phi_{kr}$ was set as $k=6$. The gain $K$ in both \eqref{Proposition 4 control law } and \eqref{Proposition 5 control law } was set as $K=1$. For the conformal navigation transformation, we iterated until $\|\bar{T}_n-\bar{T}_{n-1}\|<1\times 10^{-13}$. 

In the first simulation, the feedback control law \eqref{Proposition 4 control law } provided in Proposition 4 is applied to the kinematic mobile robot \eqref{kinematic}. The workspace is set up with three star-shaped internal obstacles: an ellipse, an inverse ellipse, a plum; and a non-star-shaped internal obstacle that cannot be handled by the homeomorphism-based transformations: the letter C. The goal point is set as
$x_d=(0,0)$, which the robot aims to converge to from 4 different
initial points, namely $x_0=(1,0.6),(-1,0.6),(1,-0.6)$ and $(0.05,-1)$. Fig. 2 and Fig. 3 depict the level sets of the navigation function and robot trajectories in the initial workspace and the transformed sphere world, respectively.
%In Figs. 2 and Figs. 3, we observe the level sets of the navigation function and robot trajectories in the initial workspace and the transformed sphere world, respectively. 
As we can observe, the designed feedback control law successfully performs the task of maintaining the robot in the initial workspace and the transformed sphere world, avoiding collisions and stabilizing it at the goal point. In this example, the number of iterations is 7 and $\|\bar{T}_{7}-\bar{T}_{6}\|=5.4675\times 10^{-14}$.

\begin{figure}[!t]\centering
	\includegraphics[width=8cm]{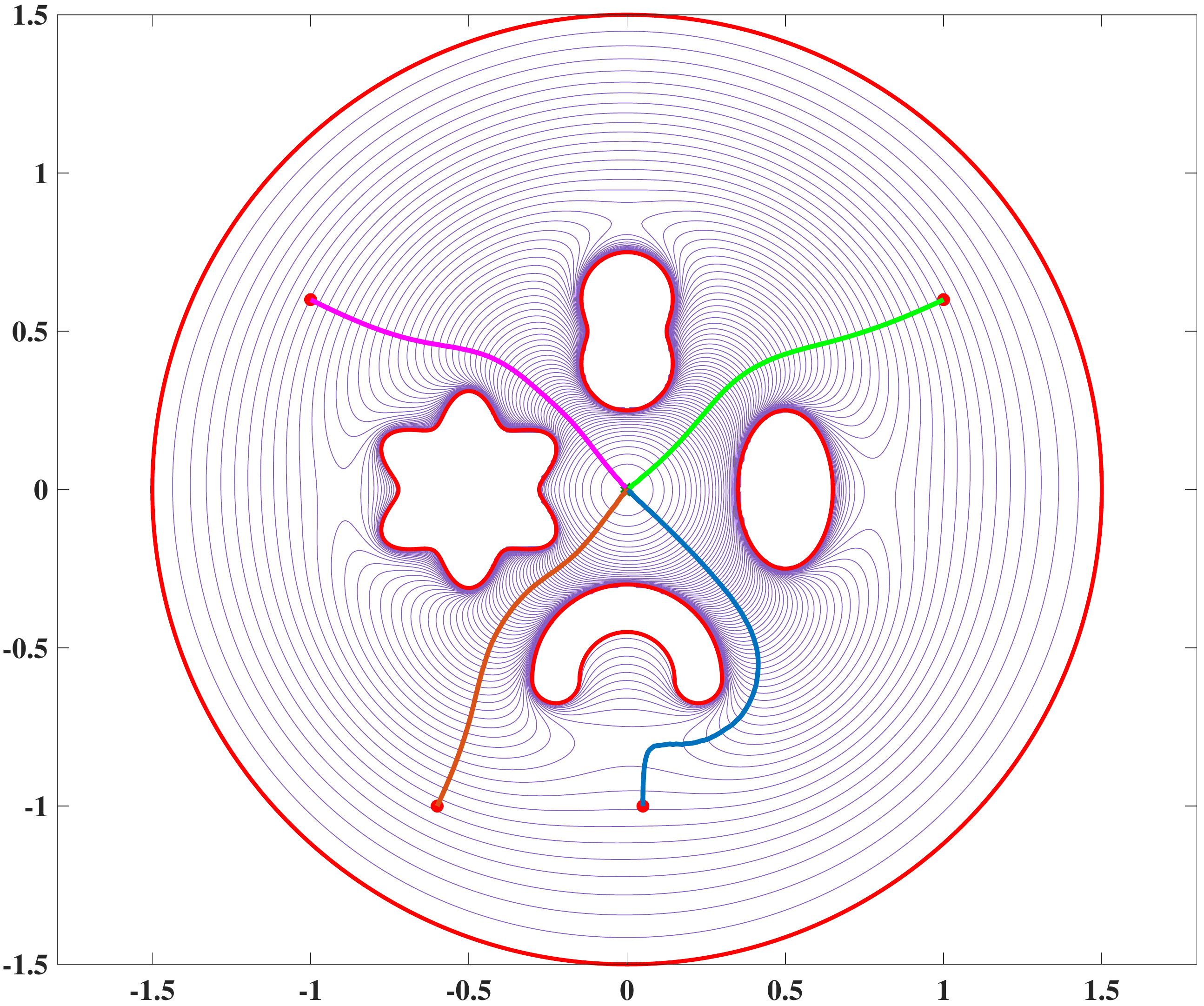}
	\caption{Trajectories of the kinematic mobile robot \eqref{kinematic} under the control law \eqref{Proposition 4 control law } for different initial points in a 2-dimensional complex workspace.}
	%Trajectory of kinematic robot \eqref{kinematic} under control law \eqref{Proposition 4 control law } in a 2D complex workspace}
	\label{Fig_2}
\end{figure}

In the second simulation, the mass of the dynamic mobile robot \eqref{dynamic} is set as $m=1kg$. The dissipative field in the control law \eqref{Proposition 5 control law } is chosen as $d(x,\dot x)=-\lambda \dot{x}$. The workspace is bounded by 5 ellipses. The initial and goal points of the robot are $x_0=(-1,0.1)$ and $x_0=(0,0)$, respectively. Fig. 4 depicts the trajectories of the dynamic robot \eqref{dynamic} under the control law \eqref{Proposition 5 control law } for different $\lambda$. The dark green trajectory corresponds to the system \eqref{dynamic} without dynamic, i.e., $\lambda=0$. The orange trajectory and the red trajectory correspond to a system with $\lambda=3.2$ and $\lambda=5$, respectively. As can be seen, all three trajectories successfully arrive and stabilize at the goal point without collision. For different constants $\lambda$, we can observe that the larger the constant $\lambda$ the closer the trajectory is to that of the system without dynamics. In this example, the number of iterations is 7 and $\|\bar{T}_{7}-\bar{T}_{6}\|=3.0234\times 10^{-14}$.

\section{Conclusion}

This paper proposes a methodology for the navigation problem in complex workspaces based on a novel spatial transformation called the conformal navigation transformation. Compared to the homeomorphism-based transformations, the proposed transformation has the advantages of requiring no tuning, existing uniquely in a given workspace, eliminating angular distortion, and being completely determined by boundary values, which are suitable for navigation problems in complex workspaces. With the application of the proposed transformation, the navigation function in the geometrically simple sphere world is pulled back to the initial workspace. The negative gradient of the pulled-back navigation function then couples with an appropriate dissipation term as a control law that enables the dynamic robotic system to converge from almost all initial points to the goal point. It is shown that this method provides solutions for both path planning and motion planning subproblems. Furthermore, an iterative method for constructing the proposed transformation is presented, which ensures fast convergence and is applicable to non-star-shaped obstacles. Finally, simulation results support the theoretical results.

%This paper proposes a methodology for the navigation problem in complex workspaces based on a space transformation called conformal navigation transformation and  a navigation function. In comparison to homeomorphic-based transformations, the proposed transformation has the advantages of requiring no tuning, existing uniquely in a given workspace, eliminating angular distortion and being completely determined by boundary values, making it suitable for navigation problems in complex workspaces. By applying the inverse transformation, the navigation function is pulled back to the workspace and its negative gradient is coupled with a suitable dissipative term to obtain a control law for the dynamic robot system that converges to the goal point from almost all initial points. Furthermore, the proposed iterative method for constructing the conformal navigation transformation ensures fast convergence and is applicable to obstacles of arbitrary shapes. In addition to the analytical guarantees of almost global robot navigation, the effectiveness of the methodology is demonstrated by numerical simulations.

\begin{figure}[!t]\centering
	\includegraphics[width=8cm]{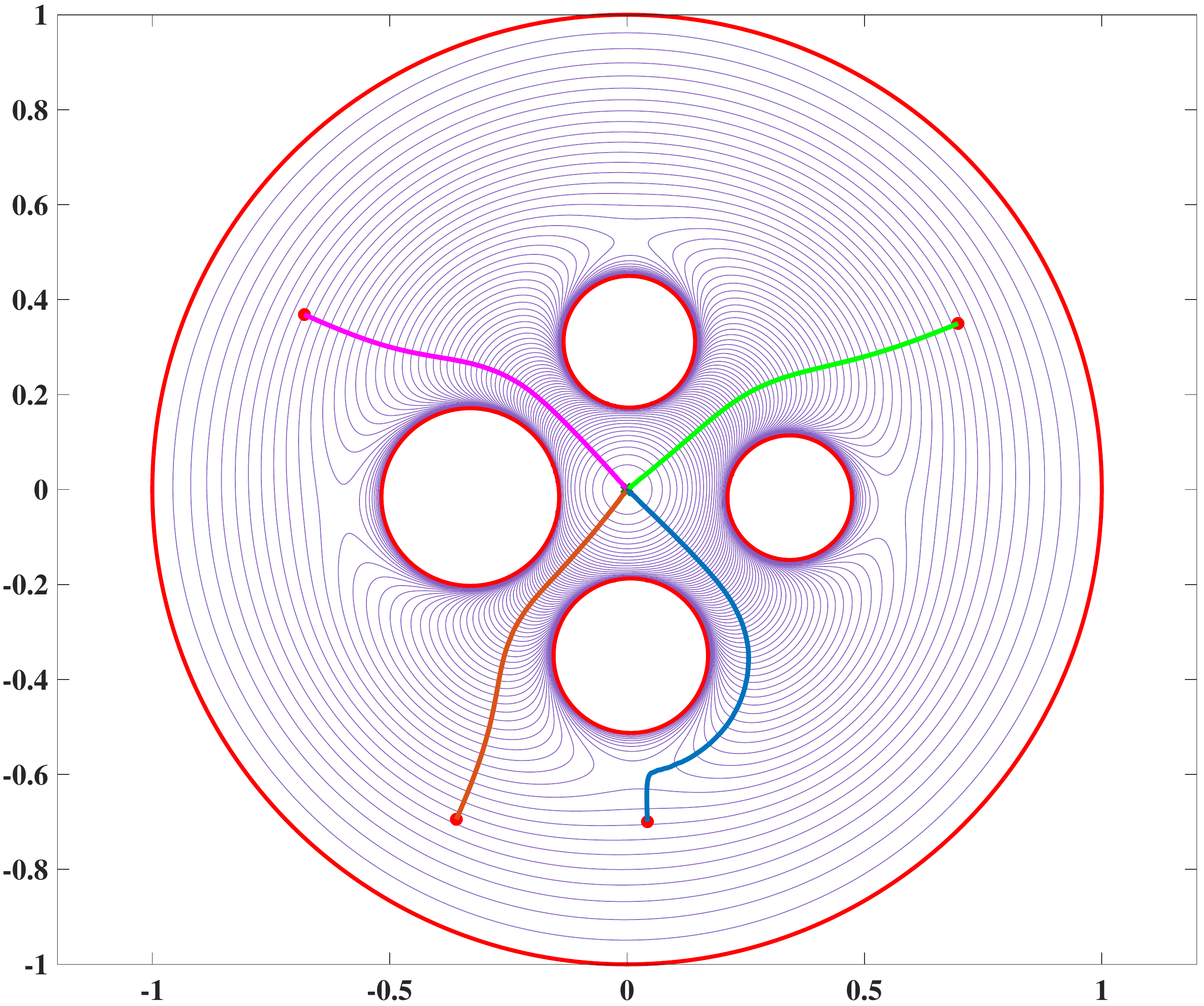}
	\caption{Trajectories of the kinematic mobile robot \eqref{kinematic} under the control law \eqref{Proposition 4 control law } for different initial points in the conformal sphere world.}
	%Trajectory of kinematic robot (4) under control law (6) in the conformal sphere world corresponding to Fig. 2}
	\label{Fig_3}
\end{figure}

Future research directions include extending the solution to account for conformal geometric structures of workspaces, constrained stabilization in complex workspaces, dynamic obstacles, multi-agent navigation problems, and more realistic scenarios with local sensing.

%\addtolength{\topmargin}{0.015in}

\section*{Appendix}%APPENDIX

\subsection{Proof of Proposition 2}

Assume that the $2$-dimensional workspace $\mathscr{N}^2$ is mapped to two sphere worlds $\mathscr{M}^2_1$ and $\mathscr{M}^2_2$ by the conformal navigation transformations $T_1$ and $T_2$, respectively. Let circles $C_1^i, i=0,1,\cdots,M$ and $C_2^i, i=0,1,\cdots,M$ be the boundaries of $\mathscr{M}^2_1$ and $\mathscr{M}^2_2$, respectively. Inverting $\mathscr{M}^2_1$ and $\mathscr{M}^2_2$ with respect to the bounding circles $C_1^i$ and $C_2^i$, respectively, and performing the inversions indefinitely, we then obtain infinitely many circles clustering to the non-dense perfect sets $P_1$ and $P_2$, respectively. 
Let $Q_1$ and $Q_2$ be the complements of $P_1$ and $P_2$ with respect to the whole plane, respectively. Since sphere worlds are special workspaces, then according to Proposition 1, there is a conformal navigation transformation $T_3$ between $\mathscr{M}^2_1$ and $\mathscr{M}^2_2$. By the Reflection Principle \cite{pommerenke1992}, $T_3$ is holomorphic in $Q_1$ and maps $Q_1$ conformally onto $Q_2$. Since $P_1$ and $P_2$ are non-dense perfect sets, $T_3$ is holomorphic on $P_1$. Hence, $T_3$ is a M{\"o}bius transformation. If the images of the prescribed three points on the circle $C_1^0$ are fixed on the circle $C_2^0$,  then $T_3$ is the identity map. Then, we have that the map $T_2\circ T_1^{-1}=T_3$ is the identity map, which completes the proof of the proposition.

\subsection{Proof of Lemma 2}

Define the function $C(z)=\frac{1}{2\pi i}\int_{\Gamma}\frac{\mu+i\upsilon}{G} \frac{d\gamma}{\gamma-z}$ %\begin{equation} C(z)=\frac{1}{2\pi i}\int_{\Gamma}\frac{\mu+i\upsilon}{G} \frac{d\gamma}{\gamma-z} \end{equation}%
for $z\notin\Gamma$. Let $C^{\pm}\big(\gamma(s)\big)=\lim_{z\to \gamma^{\pm}(s)}C(z)$. According to Sokhotski-Plemelj theorem \cite{gakhov1990}, we have that \begin{equation}\begin{split} G(s)&C^{\pm}\big(\gamma(s)\big)=\pm\frac{1}{2}\Big(\mu(s)+i\upsilon(s) \\&+\int_0^{2\pi}\frac{1}{i\pi}\frac{G(s)}{G(t)}\frac{\dot{\gamma(t)}}{\gamma(t)-\gamma(s)} \big(\mu(t)+i\upsilon(t)\big) dt\Big).\end{split}\label{Sokhotski-Plemelj theorem}\end{equation}
%Let $$\mathbf H\mu=\int_0^{2\pi}\Re\big(\frac{1}{\pi}\frac{G(s)}{G(t)}\frac{\dot{\gamma(t)}}{\gamma(t)-\gamma(s)}\big)\mu(t)dt$$ and $$\mathbf R\mu=\int_0^{2\pi}\Im\big(\frac{1}{\pi}\frac{G(s)}{G(t)}\frac{\dot{\gamma(t)}}{\gamma(t)-\gamma(s)}\big)\mu(t)dt$$
Define $\mathbf H\mu=\int_0^{2\pi}H\mu(t)dt$, $\mathbf R\upsilon=\int_0^{2\pi}R\upsilon(t)dt$, where $H=\Re\big(\frac{1}{\pi}\frac{G(s)}{G(t)}\frac{\dot{\gamma(t)}}{\gamma(t)-\gamma(s)}\big)$, $R=\Im\big(\frac{1}{\pi}\frac{G(s)}{G(t)}\frac{\dot{\gamma(t)}}{\gamma(t)-\gamma(s)}\big)$. From \eqref{Sokhotski-Plemelj theorem}, we get $\Re \big(G(s)C^{\pm}(\gamma(s))\big)=\frac{1}{2}(\pm \upsilon+\mathbf R\upsilon+\mathbf H\mu)$, $\Im \big(G(s)C^{\pm}(\gamma(s))\big)=\frac{1}{2}(\pm \mu+\mathbf R\mu+\mathbf H\upsilon)$, and the jump relation $G(s)C^+\big(\gamma(s)\big)-G(s)C^-\big(\gamma(s)\big)=\mu(s)+i\upsilon(s)$.
Let $f(z)=C(z)$. According to the jump relation of boundary values, $f(z)$ has the boundary values $G(s)f^+(\gamma(s))=G(s)C^+(\gamma(s))=G(s)C^-(\gamma(s))+\mu(s)+i\upsilon(s)$, where $f^+\big(\gamma(s)\big)=\lim_{z\to \gamma^+(s)}f(z)$. Since $f(z)$ is a holomorphic function on $U^+$, we obtain that $f(z)=C(z)=0$ for $z\in U^-$. This implies $\Im \big(G(s)C^-(\gamma(s))\big)=0$, which is equivalent to \eqref{integral equation}. Since $\Im \big(G(s)C^-(\gamma(s))\big)=0$, we have $G(s)C^-(\gamma(s))=\Re \big(G(s)C^-(\gamma(s))\big)=\frac{1}{2}(- \upsilon+\mathbf R\upsilon+\mathbf H\mu)$. Hence, the boundary values $G(s)f^+(\gamma(s))=\frac{1}{2}(- \upsilon+\mathbf R\upsilon+\mathbf H\mu)+\mu(s)+i\upsilon(s)$. This implies $c=[\mathbf H \mu-(\mathbf I-\mathbf R)\upsilon]/2$. Since $\frac{1}{2\pi}\arg{(G(s))}|_0^{2\pi}=\frac{1}{2\pi}\arg{(\gamma (s)-z_c)}|_0^{2\pi}>0$, then according to the Theorem 9 in \cite{WEGMANN2005},  \eqref{integral simple} has a unique solution, which completes the proof.

\begin{figure}[!t]\centering
	\includegraphics[width=8cm]{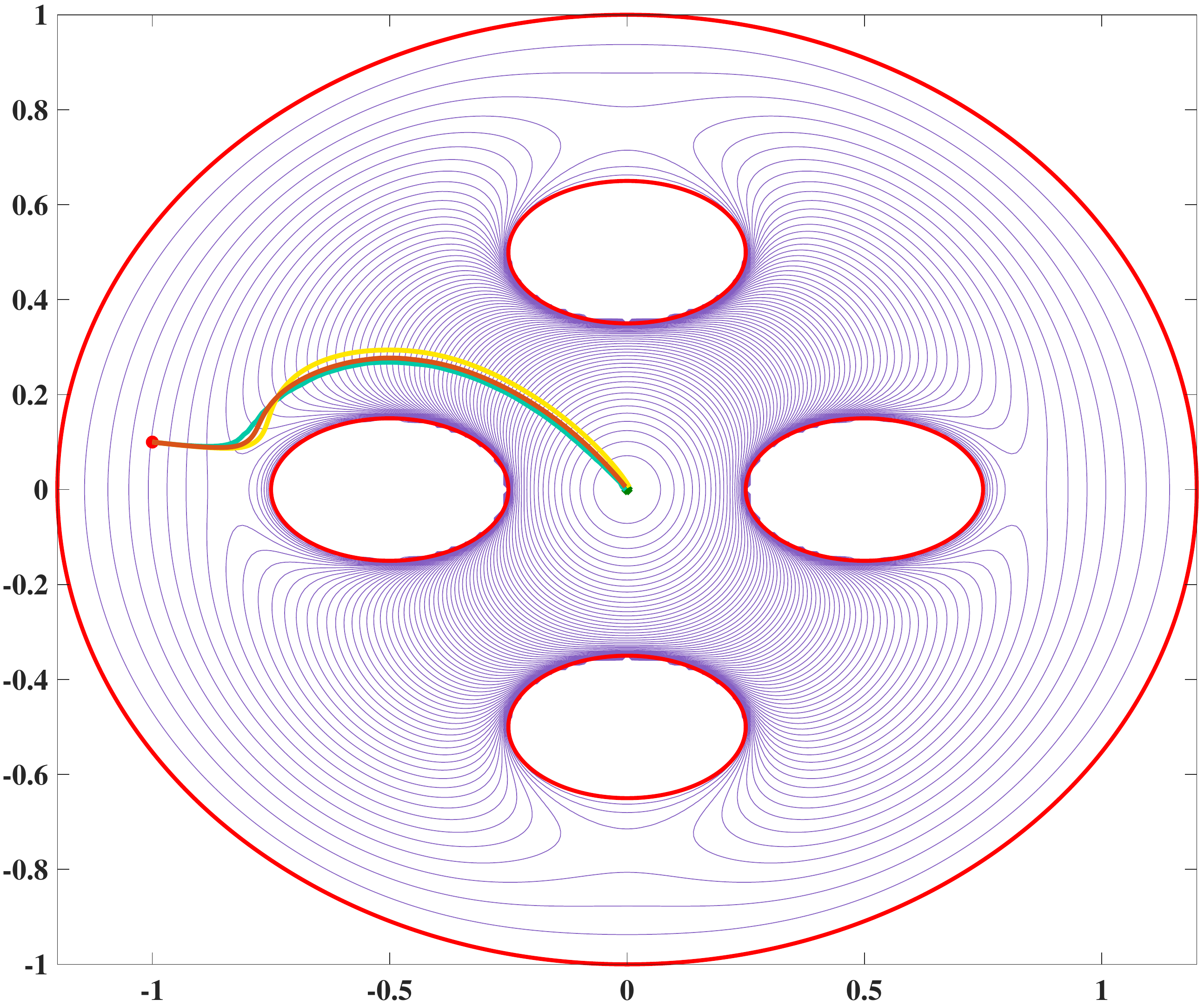}
	\caption{Trajectories of the dynamics robot  \eqref{dynamic} under control law \eqref{Proposition 5 control law } for different $\lambda$ in a 2-dimensional workspace with elliptical boundaries.}
	\label{Fig_4}
\end{figure}

\bibliographystyle{IEEEtran}
\bibliography{Bibliography/IEEEabrv,Bibliography/Mybib}\ %IEEEabrv instead of IEEEfull

\addtolength{\textheight}{-12cm}   % This command serves to balance the column lengths

%%%%%%%%%%%%%%%%%%%%%%%%%%%%%%%%%%%%%%%%%%%%%%%%%%%%%%%%%%%%%%%%%%%%%%%%%%%%%%%%

%%%%%%%%%%%%%%%%%%%%%%%%%%%%%%%%%%%%%%%%%%%%%%%%%%%%%%%%%%%%%%%%%%%%%%%%%%%%%%%%

%%%%%%%%%%%%%%%%%%%%%%%%%%%%%%%%%%%%%%%%%%%%%%%%%%%%%%%%%%%%%%%%%%%%%%%%%%%%%%%%

%%%%%%%%%%%%%%%%%%%%%%%%%%%%%%%%%%%%%%%%%%%%%%%%%%%%%%%%%%%%%%%%%%%%%%%%%%%%%%%%

\end{document}